\documentclass[11pt]{article}
\usepackage{eacl2009}
\usepackage{times}
\usepackage{graphicx}
\usepackage{latexsym}
\title{Language Diversity across the Consonant Inventories:\\ A Study in the Framework of Complex Networks}

\author{Monojit Choudhury \\ Microsoft Research India, Bangalore, India -- 560080 \\Email: {\tt 
monojitc@microsoft.com}
\AND 
Animesh Mukherjee, Anupam Basu \and Niloy Ganguly\\
Indian Institute of Technology, Kharagpur, India -- 721302
\AND
Ashish Garg \and Vaibhav Jalan\\ 
Malaviya National Institute of Technology, Jaipur, India -- 302017
}

\date{}

\begin{document}
\maketitle
\begin{abstract}
In this paper, we attempt to explain the emergence of the linguistic diversity that exists across the consonant 
inventories of some of the major language families of the world through a complex network based growth model. There is 
only a single parameter for this model that is meant to introduce a small amount of randomness in the otherwise 
preferential attachment based growth process. The experiments with this model parameter indicates that the choice of 
consonants among the languages within a family are far more preferential than it is across the families. The 
implications of this result are twofold -- (a) there is an innate preference of the speakers towards acquiring certain 
linguistic structures over others and (b) shared ancestry propels the stronger preferential connection between the 
languages within a family than across them. Furthermore, our observations indicate that this parameter might bear a 
correlation with the period of existence of the language families under investigation. 
\end{abstract}

\section{Introduction}

In one of their seminal papers~\cite{Hauser:02}, Noam Chomsky and his co-authors remarked that if a Martian ever 
graced our planet then it would be awe-struck by the unique ability of the humans to communicate among themselves 
through the medium of language. However, if our Martian naturalist were meticulous then it might also note the 
surprising co-existence of 6700 such mutually unintelligible languages across the world. Till date, the terrestrial 
scientists have no definitive answer as to why this linguistic diversity exists~\cite{Pinker:94}. Previous work in the 
area of language evolution have tried to explain the emergence of this diversity through two different background 
models. The first one among these assumes that there is a set of predefined language configurations and the movement 
of a particular language on this landscape is no more than a random walk~\cite{Tomlin:86,Dryer:92}. The second line of 
research attempts to relate the ecological, cultural and demographic parameters with the linguistic parameters 
responsible for this diversity (see for 
reference~\cite{Arita:96,Sasaki:97,Kirby:98,Livingstone:99,Nettle:99,Fought:04}). 

From the above studies it turns out that linguistic diversity is an outcome of the language dynamics in terms of its 
evolution, acquisition and change. Like any physical system, the dynamics of a linguistic system can also be viewed 
from three levels~\cite{Arhem:04}. On one extreme, it is a collection of utterances that are produced and perceived by 
the speakers of a linguistic community; this is analogous to the {\em microscopic} view of a thermodynamic system. On 
the other extreme, it can be expressed by a set of grammar rules and a lexicon; this is analogous to the {\em 
macroscopic} view. Sandwiched between these two extremes, one can also conceive a {\em mesoscopic} view of language, 
where linguistic entities such as phonemes, words or letters are the basic units and grammar is an emergent property 
of the interactions among these units. In the recent years, complex networks have proved to be an extremely suitable 
framework for modeling and studying the structure and dynamics of linguistic systems from a mesoscopic prespective 
(see~\cite{Cancho:01,Mendes:01,Cancho:04,Sole:05} for references).   

In this work, we attempt to investigate the diversity that exists across the consonant inventories of the world's 
languages through an evolutionary framework based on network growth. Along the lines of the study presented 
in~\cite{acl:06}, we model the structure of the inventories through a {\em bipartite} network, which has two different 
sets of nodes, one labeled by the languages and the other by the consonants. Edges run in between these two sets 
depending on whether a particular consonant is found in a particular language. This network is termed as the {\bf 
P}honeme--{\bf La}nguage {\bf Net}work or {\bf PlaNet} in~\cite{acl:06}. We construct five such networks that 
respectively represent the consonant inventories belonging to the five major language families namely, the 
Indo-European (IE-PlaNet), the Afro-Asiatic (AA-PlaNet), the Niger-Congo (NC-PlaNet), the Austronesian (AN-PlaNet) and 
the Sino-Tibetan (ST-PlaNet).

The emergence of the distribution of occurrence of the consonants across the languages of a family can be explained 
through a growth model for the PlaNet representing the family. We employ the {\em preferential attachment} based 
growth model introduced in~\cite{acl:06} and later analytically solved in~\cite{Peruani:07} to explain this emergence 
for each of the five families. The model involves a single parameter that is essentially meant to introduce randomness 
in the otherwise predominantly preferential growth process. We observe that the families are significantly different 
from one another in terms of this parameter value. We further observe that if we combine the inventories for all the 
families together and then attempt to fit this new data with our model, the value of the parameter is significantly 
different from that of the individual families. This indicates that the dynamics within the families is quite 
different from that across them. There are possibly two factors that regulate this dynamics: the innate preference of 
the speakers towards acquiring certain linguistic structures over others and shared ancestry of the languages within a 
family. Finally, we present a brief evolutionary history of the five families and point to a possible correlation 
between their age and the model parameter. 

The rest of the paper is laid out as follows. Section~\ref{def} states the definition of PlaNet, briefly describes the 
data source and outlines the construction procedure for the five networks.  In section~\ref{synth} we review the 
growth model for the networks. In the next section, we present the experiments with the model parameter and the 
results obtained thereby, for each of the five families. We further explain the significance of each of these results 
in the same section. We conclude in section~\ref{conc} by summarizing our contributions, pointing out some of the 
implications of the current work and indicating the possible future directions.    

\section{Definition and Construction of the Networks}\label{def}

In this section, we revisit the definition of PlaNet, discuss briefly about the data source, and explain how we 
constructed the networks for each of the families. 

\subsection{Definition of PlaNet}

\begin{figure}
\begin{center}
\includegraphics[width=1.8in]{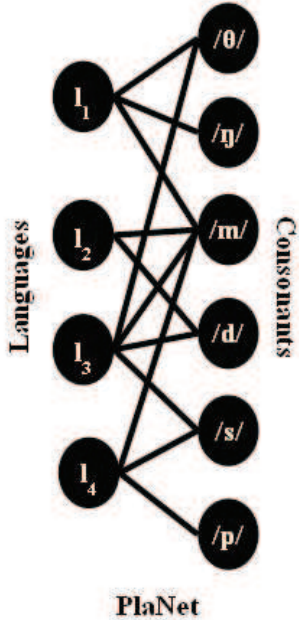}
\caption{Illustration of the nodes and edges of PlaNet.}
\label{pl}
\end{center}
\end{figure}

PlaNet is a bipartite graph $G$~=~$\langle$~$V_L$,$V_C$,$E_{pl}$~$\rangle$ consisting of two sets of nodes namely, 
$V_L$ (labeled by the languages) and $V_C$ (labeled by the consonants); $E_{pl}$ is the set of edges running between 
$V_L$ and $V_C$. There is an edge $e \in E_{pl}$ from a node $v_l \in$ $V_L$ to a node $v_c \in$ $V_C$ iff the 
consonant $c$ is present in the inventory of the language $l$. Figure~\ref{pl} illustrates the nodes and edges of 
PlaNet.

\subsection{Data Source}

We use the UCLA Phonological Segment Inventory Database (UPSID)~\cite{Maddieson:84} as the source of data for this 
work. The choice of this database is motivated by a large number of typological 
studies~\cite{Lindblom:88,Ladefoged:96,Boer:00,Hinskens:03} that have been carried out on it by the earlier 
researchers. There are 317 languages in the database with 541 consonants found across them. From these data we 
manually sort the languages into five groups representing the five families. Note that we included a language in any 
group if and only if we could find a direct evidence of its presence in the corresponding family. A brief description 
of each of these groups and languages found within them are listed below.\\
{\bf Indo-European:} This family includes most of the major languages of Europe and south, central and south-west 
Asia. Currently, it has around 3 billion native speakers, which is largest among all the recognized families of 
languages in the world. The total number of languages appearing in this family is 449. The earliest evidences of the 
Indo-European languages have been found to date 4000 years back.\\ 
{\em Languages} -- Albanian, Lithuanian, Breton, Irish, German, Norwegian, Greek, Bengali, Hindi-Urdu, Kashmiri, 
Sinhalese, Farsi, Kurdish, Pashto, French, Romanian, Spanish, Russian, Bulgarian.\\
{\bf Afro-Asiatic:} Afro-Asiatic languages have about 200 million native speakers spread over north, east, west, 
central and south-west Africa. This family is divided into five subgroups with a total of 375 languages. The 
proto-language of this family began to diverge into separate branches approximately 6000 years ago.\\  
{\em Languages} -- Shilha, Margi, Angas, Dera, Hausa, Kanakuru, Ngizim, Awiya, Somali, Iraqw, Dizi, Kefa, Kullo, 
Hamer, Arabic, Amharic, Socotri.\\
{\bf Niger-Congo:} Majority of the languages that belong to this family are found in the sub-Saharan parts of Africa. 
The number of native speakers is around 300 million and the total number of languages is 1514. This family descends 
from a proto-language, which dates back 5000 years.\\ 
{\em Languages} -- Diola, Temne, Wolof, Akan, Amo, Bariba, Beembe, Birom, Cham, Dagbani, Doayo, Efik, Ga, Gbeya, Igbo, 
Ik, Koma, Lelemi, Senadi, Tampulma, Tarok, Teke, Zande, Zulu, Kadugli, Moro, Bisa, Dan, Bambara, Kpelle.\\
{\bf Austronesian:} The languages of the Austronesian family are widely dispersed throughout the islands of south-east 
Asia and the Pacific. There are 1268 languages in this family, which are spoken by a population of 6 million native 
speakers. Around 4000 years back it separated out from its ancestral branch.\\ 
{\em Languages} -- Rukai, Tsou, Hawaiian, Iai, Adzera, Kaliai, Roro, Malagasy, Chamorro, Tagalog, Batak, Javanese.\\ 
{\bf Sino-Tibetan:} Most of the languages in this family are distributed over the entire east Asia. With a population 
of around 2 billion native speakers it ranks second after Indo-European. The total number of languages in this family 
is 403. Some of the first evidences of this family can be traced 6000 years back.\\
{\em Languages} -- Hakka, Mandarin, Taishan, Jingpho, Ao, Karen, Burmese, Lahu, Dafla.

\subsection{Construction of the Networks}

We use the consonant inventories of the languages enlisted above to construct the five bipartite networks -- 
IE-PlaNet, AA-PlaNet, NC-PlaNet, AN-PlaNet and ST-PlaNet. The number of nodes and edges in each of these networks are 
noted in Table~\ref{tab0}.

\begin{table}\centering
\begin{tabular}{llll}
\hline
\vbox to1.88ex{\vspace{1pt}\vfil\hbox to10ex{\hfil Networks\hfil}} & 
\vbox to1.88ex{\vspace{1pt}\vfil\hbox to7ex{\hfil $|V_L|$\hfil}} & 
\vbox to1.88ex{\vspace{1pt}\vfil\hbox to7ex{\hfil $|V_C|$\hfil}} & 
\vbox to1.88ex{\vspace{1pt}\vfil\hbox to7ex{\hfil $|E_{pl}|$\hfil}} \\

\hline
\hline
\vbox to1.88ex{\vspace{1pt}\vfil\hbox to10ex{\hfil IE-PlaNet\hfil}} & 
\vbox to1.88ex{\vspace{1pt}\vfil\hbox to7ex{\hfil 19\hfil}} & 
\vbox to1.88ex{\vspace{1pt}\vfil\hbox to7ex{\hfil 148\hfil}} & 
\vbox to1.88ex{\vspace{1pt}\vfil\hbox to7ex{\hfil 534\hfil}} \\

\vbox to1.88ex{\vspace{1pt}\vfil\hbox to10ex{\hfil AA-PlaNet\hfil}} & 
\vbox to1.88ex{\vspace{1pt}\vfil\hbox to7ex{\hfil 17\hfil}} & 
\vbox to1.88ex{\vspace{1pt}\vfil\hbox to7ex{\hfil 123\hfil}} & 
\vbox to1.88ex{\vspace{1pt}\vfil\hbox to7ex{\hfil 453\hfil}} \\

\vbox to1.88ex{\vspace{1pt}\vfil\hbox to10ex{\hfil NC-PlaNet\hfil}} & 
\vbox to1.88ex{\vspace{1pt}\vfil\hbox to7ex{\hfil 30\hfil}} & 
\vbox to1.88ex{\vspace{1pt}\vfil\hbox to7ex{\hfil 135\hfil}} & 
\vbox to1.88ex{\vspace{1pt}\vfil\hbox to7ex{\hfil 692\hfil}} \\

\vbox to1.88ex{\vspace{1pt}\vfil\hbox to10ex{\hfil AN-PlaNet\hfil}} & 
\vbox to1.88ex{\vspace{1pt}\vfil\hbox to7ex{\hfil 12\hfil}} & 
\vbox to1.88ex{\vspace{1pt}\vfil\hbox to7ex{\hfil 82\hfil}} & 
\vbox to1.88ex{\vspace{1pt}\vfil\hbox to7ex{\hfil 221\hfil}} \\

\vbox to1.88ex{\vspace{1pt}\vfil\hbox to10ex{\hfil ST-PlaNet\hfil}} & 
\vbox to1.88ex{\vspace{1pt}\vfil\hbox to7ex{\hfil 9\hfil}} & 
\vbox to1.88ex{\vspace{1pt}\vfil\hbox to7ex{\hfil 71\hfil}} & 
\vbox to1.88ex{\vspace{1pt}\vfil\hbox to7ex{\hfil 201\hfil}} \\

\cline{1-4}
\end{tabular}
\caption{Number of nodes and edges in the five bipartite networks corresponding to the five families.}
\label{tab0}
\end{table}
 
\section{The Growth Model for the Networks}\label{synth}

As mentioned earlier, we employ the growth model introduced in~\cite{acl:06} and later (approximately) solved 
in~\cite{Peruani:07} to explain the emergence of the {\em degree distribution} of the consonant nodes for the five 
bipartite networks. For the purpose of readability, we briefly summarize the idea below.       

\noindent{\bf Degree Distribution:} The degree of a node $v$, denoted by $k$, is the number of edges incident on $v$. 
The degree distribution is the fraction of nodes $p_k$ that have a degree equal to $k$~\cite{Newman:03}. The 
cumulative degree distribution $P_k$ is the fraction of nodes having degree greater than or equal to $k$. Therefore, 
if there are $N$ nodes in a network then,
\begin{equation}\label{eq1}
P_k = \sum_{k=k'}^{N} p_{k'}
\end{equation}  

\noindent{\bf Model Description:} The model assumes that the size of the consonant inventories (i.e., the degree of 
the language nodes in PlaNet) are known {\em a priori}. 

Let the degree of a language node $L_i \in$ $V_L$ be denoted by $d_i$ (i.e., $d_i$ refers to the inventory size of the 
language $L_i$ in UPSID). The consonant nodes in $V_C$ are assumed to be unlabeled, i.e, they are not marked by the 
articulatory/acoustic features (see~\cite{Trub:30} for further reference) that characterize them. The nodes $L_1$ 
through $L_{317}$ are sorted in the ascending order of their degrees. At each time step a node $L_j$, chosen in order, 
preferentially gets connected to $d_j$ {\em distinct} nodes (call each such node $C$) of the set $V_C$. The 
probability $Pr(C)$ with which the node $L_j$ gets connected to the node $C$ is given by, 
\begin{equation}\label{pref}
Pr(C) = \frac{k + \epsilon}{\sum_{\forall C^{'}} (k^{'} + \epsilon)}
\end{equation} where, $k$ is the current degree of the node $C$, $C^{'}$ represents the nodes in $V_C$ that are not 
already connected to $L_j$ and $\epsilon$ is the model parameter that is meant to introduce a small amount of 
randomness into the growth process. The above steps are repeated until all the language nodes $L_j \in V_L$ get 
connected to $d_j$ consonant nodes. 

Peruani \bgroup et al.\egroup~\shortcite{Peruani:07} have shown that after some simplifications one can exactly solve 
this model analytically. Let the average consonant inventory size be denoted by $\mu$ and the number of consonant 
nodes be {\em N}. The simplified model assumes that at each time step $t$ a language node gets attached to $\mu$ 
consonant nodes, following the distribution $Pr(C)$. Under the above assumptions, the degree distribution $p_{k,t}$ 
for the consonant nodes, obtained by solving the model, is a $\beta$-distribution as follows
\begin{equation}\label{eq3}
p_{k,t} \simeq A\left(\frac{k}{t}\right)^{\epsilon-1}\left(1-\frac{k}{t}\right)^{\frac{N\epsilon}{\mu}-\epsilon-1}
\end{equation} 
where $A$ is a constant term. Using equations~\ref{eq1} and~\ref{eq3} one can easily compute the value of $P_{k,t}$. 
There is a subtle point that needs a mention here. The concept of a {\em time step} is very crucial for a growing 
network. It might refer to the addition of an edge or a node to the network. While these two concepts coincide when 
every new node has exactly one edge, there are obvious differences when the new node has degree greater than one. The 
analysis presented in Peruani et al.~\shortcite{Peruani:07} holds good for the case when only one edge is added per 
time step. However, if the degree of the new node being introduced to the system is much less than $N$, then 
Eq.~\ref{eq3} is a good approximation of the emergent degree distribution for the case when a node with more than one 
edge is added per time step. Therefore, the experiments presented in the next section attempt to fit the degree 
distribution of the real networks with Eq.~\ref{eq3} by tuning the parameter $\epsilon$.

\section{Experiments and Results}\label{expt}

\begin{figure*}
\begin{center}
\includegraphics[width=6in]{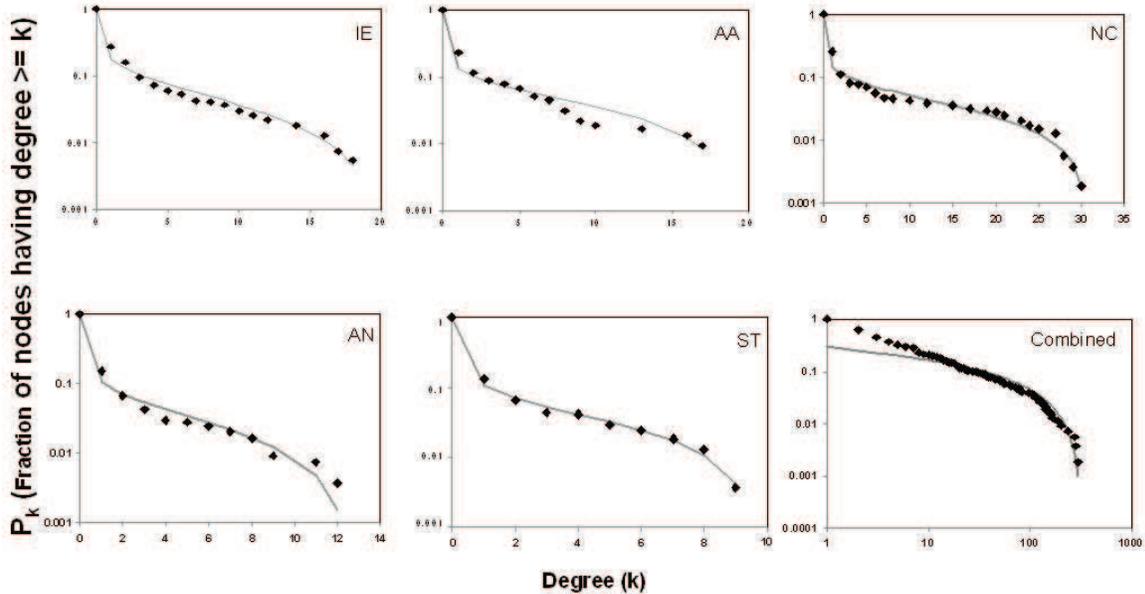}
\caption{The degree distribution of the different real networks along with the fits obtained from the equation for the 
optimal values of $\epsilon$. Black dots indicate plots for the real networks while the grey lines represent the 
curves obtained from the equation. For each of the families, the y-axis is in log-scale. The last figure showing the 
curve for the Combined-PlaNet is in doubly logarithmic scale.}
\label{fig2}
\end{center}
\end{figure*}

In this section, we attempt to fit the degree distribution of the five empirical networks with the expression for 
$P_{k,t}$ described in the previous section. For all the experiments we set $N = 541$, $t =$ number of languages in 
the family under investigation and $\mu =$ average degree of the language nodes of the PlaNet representing the family 
under investigation, that is, the average inventory size for the family. Therefore, given the value of $k$ we can 
compute $p_{k,t}$ and consequently, $P_{k,t}$, if $\epsilon$ is known. We vary the value of $\epsilon$ such that the 
logarithmic standard error ($LSE$) between the degree distribution of the real network and the equation is least. 
$LSE$ is defined as the sum of the square of the difference between the logarithm of the ordinate pairs (say $y$ and 
$y^{'}$) for which the abscissas are equal. In other words $LSE = (\log{y}-\log{y'})^2$. The best fits obtained for 
each of the five networks are shown in Figure~\ref{fig2}. The values of $\epsilon$ and the corresponding least $LSE$ 
for each of them are noted in Table~\ref{tab1}. Note that since we varied $\epsilon$ in steps of 0.005 during the 
experiments, some of the differences in the values of $\epsilon$ in Table~\ref{tab1} are not significant. 
Nevertheless, there are certain observations described below that are statistically significant according to the above 
experiments.  

\begin{table}\centering
\resizebox{!}{0.48in}{
\begin{tabular}{lll}
\hline

\vbox to1.88ex{\vspace{1pt}\vfil\hbox to12.80ex{\hfil Network\hfil}} & 
\vbox to1.88ex{\vspace{1pt}\vfil\hbox to26.20ex{\hfil $\epsilon$ for least $LSE$\hfil}} & 
\vbox to1.88ex{\vspace{1pt}\vfil\hbox to13.60ex{\hfil Value of $LSE$\hfil}} \\

\hline
\hline
\vbox to1.88ex{\vspace{1pt}\vfil\hbox to12.80ex{\hfil IE-PlaNet\hfil}} & 
\vbox to1.88ex{\vspace{1pt}\vfil\hbox to26.20ex{\hfil 0.055\hfil}} & 
\vbox to1.88ex{\vspace{1pt}\vfil\hbox to13.60ex{\hfil 0.16\hfil}} \\

\vbox to1.88ex{\vspace{1pt}\vfil\hbox to12.80ex{\hfil AA-PlaNet\hfil}} & 
\vbox to1.88ex{\vspace{1pt}\vfil\hbox to26.20ex{\hfil 0.040\hfil}} & 
\vbox to1.88ex{\vspace{1pt}\vfil\hbox to13.60ex{\hfil 0.24\hfil}} \\

\vbox to1.88ex{\vspace{1pt}\vfil\hbox to12.80ex{\hfil NC-PlaNet\hfil}} & 
\vbox to1.88ex{\vspace{1pt}\vfil\hbox to26.20ex{\hfil 0.035\hfil}} & 
\vbox to1.88ex{\vspace{1pt}\vfil\hbox to13.60ex{\hfil 0.19\hfil}} \\

\vbox to1.88ex{\vspace{1pt}\vfil\hbox to12.80ex{\hfil AN-PlaNet\hfil}} & 
\vbox to1.88ex{\vspace{1pt}\vfil\hbox to26.20ex{\hfil 0.030\hfil}} & 
\vbox to1.88ex{\vspace{1pt}\vfil\hbox to13.60ex{\hfil 0.17\hfil}} \\

\vbox to1.88ex{\vspace{1pt}\vfil\hbox to12.80ex{\hfil ST-PlaNet\hfil}} & 
\vbox to1.88ex{\vspace{1pt}\vfil\hbox to26.20ex{\hfil 0.035\hfil}} & 
\vbox to1.88ex{\vspace{1pt}\vfil\hbox to13.60ex{\hfil 0.03\hfil}} \\

\vbox to1.88ex{\vspace{1pt}\vfil\hbox to12.80ex{\hfil Combined-PlaNet\hfil}} & 
\vbox to1.88ex{\vspace{1pt}\vfil\hbox to26.20ex{\hfil 0.070\hfil}} & 
\vbox to1.88ex{\vspace{1pt}\vfil\hbox to13.60ex{\hfil 1.47\hfil}} \\

\hline
\end{tabular}}
\caption{The values of $\epsilon$ and the least $LSE$ for the different networks. Combined-PlaNet refers to the 
network constructed after mixing all the languages from all the families. For all the experiments}
\label{tab1}
\end{table}

\noindent{\bf Observation I:} The very low value of the parameter $\epsilon$ indicates that the choice of consonants 
within the languages of a family is strongly preferential. This innate preference for acquiring a particular set of 
consonants tends to grow over the linguistic generations~\cite{Blevins:04}. In this context, $\epsilon$ may be thought 
of as modeling the (accidental) errors or drifts that can occur during language transmission. The fact that the values 
of $\epsilon$ across the four major language families, namely Afro-Asiatic,Niger-Congo, Sino-Tibetan and Austronesian, 
are comparable indicates that the rate of error propagation is a universal factor that is largely constant across the 
families. The value of $\epsilon$ for IE-PlaNet is slightly higher than the other four families, which might be an 
effect of higher diversification within the family due to geographical or socio-political factors. Nevertheless, it is 
still smaller than the $\epsilon$ of the Combined-PlaNet.

The optimal $\epsilon$ obtained for Combined-PlaNet is higher than that of all the families (see Table~\ref{tab1}), 
though it is comparable to the Indo-European PlaNet. This points to the fact that the choice of consonants within the 
languages of a family is far more preferential than it is across the families; this fact is possibly an outcome of 
shared ancestry. In other words, the inventories of genetically related languages are similar (i.e., they share a lot 
of consonants) because they have evolved from the same parent language through a series of linguistic changes, and the 
chances that they use a large number of consonants used by the parent language is naturally high. 

\noindent{\bf Observation II:} We observe a very interesting relationship between the approximate age of the language 
family and the values of $\epsilon$ obtained in each case (see Table~\ref{tab4}). The only anomaly is the 
Indo-European branch, which possibly indicates that this might be much older than it is believed to be. In fact, a 
recent study~\cite{Balter:03} has shown that the age of this family dates back to 8000 years. If this last argument is 
assumed to be true then the values of $\epsilon$ have a one-to-one correspondence with the approximate period of 
existence of the language families. As a matter of fact, this correlation can be intuitively justified -- higher is 
the period of existence of a family higher are the chances of its diversification into smaller subgroups, which in 
turn increases the chances of transmission errors and hence, the values of $\epsilon$ comes out to be more for the 
older families. It should be noted that the difference between the values of $\epsilon$ for the language families are 
not statistically significant. Therefore, the aforementioned observation should be interpreted only as an interesting 
possibility; more experimentation is required for making any stronger claim.

\subsection{Control Experiment}
How could one be sure that the aforementioned observations are not an obvious outcome of the construction of the 
PlaNet or some spurious correlations? To this end, we conduct a control experiment where a set of inventories is 
randomly selected from UPSID to represent a family. The number of languages chosen is same as that of the PlaNets of 
the various language families. We observe that the average value of $\epsilon$ for these randomly constructed PlaNets 
is 0.068, which, as one would expect, is close to that of the Combined-PlaNet. This reinforces the fact that the 
inherent proximity among the languages of a real family is not a consequence ``by chance".

\subsection{Correlation between Families}
Another way to verify the above observation is to estimate the correlation between the frequency of occurrence of the 
consonants for the different language family pairs (i.e., how the frequencies of the consonants /p/, /t/, /k/, /m/, 
/n/ {\dots} are correlated across the different families). Table~\ref{tab3} notes the value of this correlation among 
the five families. The values in Table~\ref{tab3} indicate that, in general, the families are very weakly correlated 
with each other, the average correlation being as low as $\sim 0.47$. 

Note that, the correlation between the Afro-Asiatic and the Niger-Congo families is high not only because they share 
the same African origin, but also due to higher chances of language contacts among their groups of speakers. On the 
other hand, the Indo-European and the Sino-Tibetan families show least correlation because it is usually believed that 
they share absolutely no genetic connections. Interestingly, similar trends are observed for the values of the 
parameter $\epsilon$. If we combine the languages of the Afro-Asiatic and the Niger-Congo families and try to fit the 
new data then $\epsilon$ turns out to be 0.035 while if we do the same for the Indo-European and the Sino-Tibetan 
families then $\epsilon$ is 0.058. For many of the other combinations the value of $\epsilon$ and the correlation 
coefficient have a one-to-one correspondence. However, there are clear exceptions also. For instance, if we combine 
the Afro-Asiatic and the Indo-European families then the value of $\epsilon$ is very low (close to 0.04) although the 
correlation between them is not very high. The reasons for these exceptions should be interesting and we plan to 
further explore this issue in future.  
        
\begin{table}\centering
\resizebox{!}{0.45in}{
\begin{tabular}{llllll}
\hline
\vbox to1.88ex{\vspace{1pt}\vfil\hbox to10.40ex{\hfil Families\hfil}} & 
\vbox to1.88ex{\vspace{1pt}\vfil\hbox to10.40ex{\hfil IE\hfil}} & 
\vbox to1.88ex{\vspace{1pt}\vfil\hbox to07.00ex{\hfil AA\hfil}} & 
\vbox to1.88ex{\vspace{1pt}\vfil\hbox to07.00ex{\hfil NC\hfil}} & 
\vbox to1.88ex{\vspace{1pt}\vfil\hbox to07.40ex{\hfil AN\hfil}} & 
\vbox to1.88ex{\vspace{1pt}\vfil\hbox to07.20ex{\hfil ST\hfil}} \\

\hline
\hline
\vbox to1.88ex{\vspace{1pt}\vfil\hbox to10.40ex{\hfil IE\hfil}} & 
\vbox to1.88ex{\vspace{1pt}\vfil\hbox to10.40ex{\hfil --\hfil}} & 
\vbox to1.88ex{\vspace{1pt}\vfil\hbox to07.00ex{\hfil 0.49\hfil}} & 
\vbox to1.88ex{\vspace{1pt}\vfil\hbox to07.00ex{\hfil 0.48\hfil}} & 
\vbox to1.88ex{\vspace{1pt}\vfil\hbox to07.40ex{\hfil 0.42\hfil}} & 
\vbox to1.88ex{\vspace{1pt}\vfil\hbox to07.20ex{\hfil 0.25\hfil}} \\

\vbox to1.88ex{\vspace{1pt}\vfil\hbox to10.40ex{\hfil AA\hfil}} & 
\vbox to1.88ex{\vspace{1pt}\vfil\hbox to10.40ex{\hfil 0.49\hfil}} & 
\vbox to1.88ex{\vspace{1pt}\vfil\hbox to07.00ex{\hfil --\hfil}} & 
\vbox to1.88ex{\vspace{1pt}\vfil\hbox to07.00ex{\hfil 0.66\hfil}} & 
\vbox to1.88ex{\vspace{1pt}\vfil\hbox to07.40ex{\hfil 0.53\hfil}} & 
\vbox to1.88ex{\vspace{1pt}\vfil\hbox to07.20ex{\hfil 0.43\hfil}} \\

\vbox to1.88ex{\vspace{1pt}\vfil\hbox to10.40ex{\hfil NC\hfil}} & 
\vbox to1.88ex{\vspace{1pt}\vfil\hbox to10.40ex{\hfil 0.48\hfil}} & 
\vbox to1.88ex{\vspace{1pt}\vfil\hbox to07.00ex{\hfil 0.66\hfil}} & 
\vbox to1.88ex{\vspace{1pt}\vfil\hbox to07.00ex{\hfil --\hfil}} & 
\vbox to1.88ex{\vspace{1pt}\vfil\hbox to07.40ex{\hfil 0.55\hfil}} & 
\vbox to1.88ex{\vspace{1pt}\vfil\hbox to07.20ex{\hfil 0.37\hfil}} \\

\vbox to1.88ex{\vspace{1pt}\vfil\hbox to10.40ex{\hfil AN\hfil}} & 
\vbox to1.88ex{\vspace{1pt}\vfil\hbox to10.40ex{\hfil 0.42\hfil}} & 
\vbox to1.88ex{\vspace{1pt}\vfil\hbox to07.00ex{\hfil 0.53\hfil}} & 
\vbox to1.88ex{\vspace{1pt}\vfil\hbox to07.00ex{\hfil 0.55\hfil}} & 
\vbox to1.88ex{\vspace{1pt}\vfil\hbox to07.40ex{\hfil --\hfil}} & 
\vbox to1.88ex{\vspace{1pt}\vfil\hbox to07.20ex{\hfil 0.50\hfil}} \\

\vbox to1.88ex{\vspace{1pt}\vfil\hbox to10.40ex{\hfil ST\hfil}} & 
\vbox to1.88ex{\vspace{1pt}\vfil\hbox to10.40ex{\hfil 0.25\hfil}} & 
\vbox to1.88ex{\vspace{1pt}\vfil\hbox to07.00ex{\hfil 0.43\hfil}} & 
\vbox to1.88ex{\vspace{1pt}\vfil\hbox to07.00ex{\hfil 0.37\hfil}} & 
\vbox to1.88ex{\vspace{1pt}\vfil\hbox to07.40ex{\hfil 0.50\hfil}} & 
\vbox to1.88ex{\vspace{1pt}\vfil\hbox to07.20ex{\hfil --\hfil}} \\

\cline{1-6}
\end{tabular}}
\caption{The Pearson's correlation between the frequency distributions obtained for the family pairs. IE: 
Indo-European, AA: Afro-Asiatic, NC: Niger-Congo, AN: Austronesian, ST: Sino-Tibetan.}
\label{tab3}
\end{table}

\begin{table}\centering
\resizebox{!}{0.55in}{
\begin{tabular}{ccc}
\hline
Families & Age (in years) & $\epsilon$ \\ 

\hline
\hline
Austronasean & 4000 & 0.030 \\

Niger-Congo & 5000 & 0.035 \\

Sino-Tibetan & 6000 & 0.035 \\

Afro-Asiatic & 6000 & 0.040 \\

Indo-European & 4000 (or 8000) & 0.055 \\ 

\hline
\end{tabular}}
\caption{Table showing the relationship between the age of a family and the value of $\epsilon$.}
\label{tab4}
\end{table}

\section{Conclusion}\label{conc}

In this paper, we presented a method of network evolution to capture the emergence of linguistic diversity that 
manifests in the five major language families of the world. The bipartite network based growth model that we proposed 
in this paper can be associated with the process of language acquisition by an individual, which largely governs the 
course of language change in a linguistic community. In the initial years of language development every child passes 
through a stage called {\em babbling} during which he/she learns to produce non-meaningful sequences of consonants and 
vowels, some of which are not even used in the language to which they are exposed~\cite{Jakob:68,Locke:83}. Clear 
preferences can be observed for learning certain sounds such as plosives and nasals, whereas fricatives and liquids 
are avoided. In fact, this hierarchy of preference during the babbling stage follows the cross-linguistic frequency 
distribution of the consonants. This innate frequency dependent preference towards certain phonemes might be because 
of phonetic reasons (i.e., for articulatory/perceptual benefits). In the current model, this innate preference gets 
captured through the process of preferential attachment. 
 
In fact, preferential attachment (PA) is a universally observed evolutionary mechanism that is known to shape several 
physical, biological and socio-economic systems~\cite{Newman:03}. This phenomenon has also been called for to explain 
various linguistic phenomena~\cite{eccs:08}. We believe that PA also provides a suitable abstraction for the mechanism 
of language acquisition. Acquisition of vocabulary and growth of the mental lexicon are few examples of PA in language 
acquisition. This work illustrates another variant of PA applied to explain the structure of consonant inventories and 
their diversification across the language families.

\end{document}